\documentclass{article}

\PassOptionsToPackage{numbers,compress}{natbib}
\usepackage[final]{neurips_2026}

\usepackage[utf8]{inputenc} 
\usepackage[T1]{fontenc}    
\usepackage{hyperref}       
\usepackage{url}            
\usepackage{booktabs}       
\usepackage{amsfonts}       
\usepackage{nicefrac}       
\usepackage{microtype}      
\usepackage{xcolor}         

\usepackage{graphicx}
\usepackage{cleveref}
\usepackage[table,xcdraw]{xcolor}
\usepackage{booktabs}
\usepackage{multirow}
\usepackage{pdflscape}
\usepackage{color}
\usepackage{tabularray}
\usepackage{comment}
\usepackage{pifont} 
\usepackage[T1]{fontenc} 
\usepackage{caption}
\usepackage{fontawesome5}

\newcommand{\cmark}{\ding{51}} 
\newcommand{\xmark}{\ding{55}} 

\title{SynDORBench: Evaluating LVLM Perceptual Robustness Under Physically Constrained Visibility Conditions}

\author{%
  Jeremy Stephen Gabriel Yee$^{1,\dagger,\textsuperscript{\faEnvelope}}$ \qquad
  Zhengkui Wang$^{1}$ \qquad
  Zhiyuan Zhang$^{2}$ \\
  \textbf{Avinash Anand}$^{1}$ \qquad
  \textbf{Timothy Liu}$^{3}$ \qquad
  \textbf{Benedict Chan}$^{1}$ \qquad
  \textbf{Aik Beng Ng}$^{3}$ \qquad
  \textbf{Simon See}$^{3}$ \qquad
  \\
  \\
  $^{1}$Singapore Institute of Technology \\
  $^{2}$Singapore Management University \\
  $^{3}$NVIDIA AI Technology Center \\
  $^{\dagger}$Project lead \qquad
  \faEnvelope\ Corresponding Author
}

\makeatletter
\providecommand{\@trackname}{Evaluations and Datasets Track}
\makeatother

\begin{document}

\maketitle

\begin{abstract}
Large vision-language models (LVLMs) have demonstrated remarkable performance on multimodal reasoning benchmarks, yet their perceptual reliability under physically constrained imaging conditions remains poorly understood. Existing evaluations predominantly assume ideal visual inputs and therefore fail to characterize how camera distance, illumination, viewpoint, and pixel density fundamentally affect semantic recoverability. We introduce SynDORBench, the first physically grounded benchmark for evaluating LVLM perceptual robustness under DORI-calibrated conditions aligned with human visual capability standards. SynDORBench comprises over 54k question--answer pairs generated through a controllable synthetic pipeline that systematically varies viewing distance, lighting, camera geometry, and action pose according to physically interpretable pixel-density regimes. To support scalable low-visibility supervision, we further propose a discernibility annotation framework that propagates human perceptual labels using mask-conditioned statistical features and ensemble learning. We evaluate 16 open-source LVLMs, a commercial LVLM baseline, and YOLO11x across human-presence classification and action recognition tasks under progressively degraded visibility conditions. Our results reveal that perceptual failure in LVLMs is strongly governed by pixel density and physical imaging constraints rather than model scale alone. Surprisingly, several compact open-source LVLMs outperform larger commercial baselines and substantially exceed YOLO11x robustness under long-range and low-light conditions. SynDORBench establishes a new benchmark paradigm for physically grounded multimodal evaluation, enabling systematic analysis of LVLM reliability under real-world perceptual constraints and direct comparison against human visibility thresholds. SynDORBench is available at: \href{https://huggingface.co/datasets/jzyee/SynDORBench}{Link} and evaluation code is available at: \href{https://github.com/jzyee/SynDORBench}{Link}

    
\end{abstract}

\section{Introduction}

Large language models (LLMs) have rapidly evolved from language understanding systems into general multimodal foundation models capable of jointly reasoning over text, images, and video. Building upon advances in multimodal representation learning, Large Vision--Language Models (LVLMs) now demonstrate impressive performance across a broad spectrum of visual reasoning and perception tasks~\cite{marafioti2025smolvlm,team2025gemma,zhu2025internvl3}. Existing benchmarks such as MMBench~\cite{liu2024mmbench}, MMMU~\cite{yue2024mmmu}, and SEED-Bench~\cite{li2023seed} show that LVLMs can achieve strong semantic understanding under high-quality visual conditions. However, these evaluations predominantly assume idealized inputs, where subjects are well-lit, visually salient, and captured at sufficiently high resolution.

In contrast, real-world visual perception is fundamentally constrained by imaging physics. In practical deployment environments such as robotics, autonomous systems, and edge intelligence, visual evidence degrades continuously as a function of camera distance, illumination, sensor characteristics, viewpoint, and environmental interference~\cite{liu2025surveillancevqa}. These factors directly determine pixel density and therefore constrain the recoverability of semantic information from the scene. Consequently, an LVLM may demonstrate strong reasoning capability under ideal imaging conditions while failing catastrophically once the available visual evidence falls below perceptual thresholds. Existing evaluations primarily measure semantic competence rather than perceptual robustness, but not how far, under what visibility conditions, or under which physical constraints these systems remain operationally reliable.


\begin{figure}[tp]
    \centering
    \includegraphics[width=0.75\columnwidth]{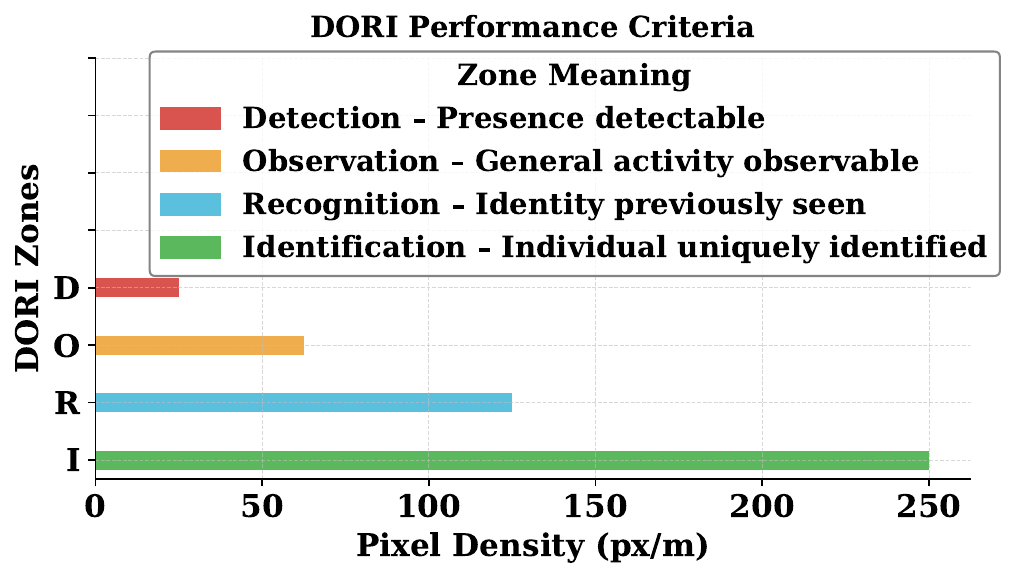}
    \caption{Minimum pixel density (px/m) required for human perceptual tasks under the BS EN 62676-4:2015 DORI standard~\cite{bsen62676-4}. The DORI framework defines physically grounded visibility thresholds for Detection, Observation, Recognition, and Identification.}
    \label{fig:dori_bar}
\end{figure}

To address this gap, we draw inspiration from the BS EN 62676-4:2015 DORI standard~\cite{bsen62676-4}, a widely adopted imaging framework that defines four perceptual visibility levels: \underline{D}etection, \underline{O}bservation, \underline{R}ecognition, and \underline{I}dentification (DORI). Each level corresponds to the minimum pixel density required for a human observer to reliably perform a visual task (Fig.~\ref{fig:dori_bar}). While DORI has long served as a physically grounded model of human visual capability, it has never been used to characterize the perceptual robustness of LVLMs. Consequently, the relationship between multimodal reasoning performance and physically measurable imaging constraints remains largely unexplored.

In this work, we introduce \textbf{SynDORBench}, the first physically grounded benchmark designed to evaluate LVLM perceptual robustness under DORI-calibrated imaging conditions. Rather than evaluating models solely on semantic correctness, SynDORBench systematically studies how multimodal perception degrades as available visual evidence deteriorates. To achieve this, we construct a controllable synthetic evaluation framework that explicitly models camera distance, illumination, viewpoint, and pixel-density regimes according to physically interpretable imaging constraints. The benchmark contains over 54k question--answer pairs spanning human-presence classification and action recognition tasks across progressively degraded visibility conditions.

Unlike existing multimodal benchmarks that rely on uncontrolled natural variation, SynDORBench enables factorized robustness analysis by isolating individual physical degradation factors while maintaining precise control over camera geometry and perceptual visibility. Furthermore, to support scalable evaluation under low-light and night-vision conditions, we introduce a discernibility annotation framework that propagates human perceptual labels through mask-conditioned statistical features and ensemble learning.

Extensive experiments across 16 open-source LVLMs, a commercial LVLM baseline, and YOLO11x reveal several important findings. First, perceptual failure in LVLMs is strongly governed by pixel density and imaging constraints rather than model scale alone. Second, modern LVLMs substantially outperform traditional detection pipelines under long-range and low-visibility conditions, suggesting that multimodal reasoning architectures exhibit greater robustness to perceptual degradation than conventional detector-based systems. Third, several compact open-source LVLMs achieve robustness comparable to or exceeding larger commercial baselines, indicating that physically grounded perceptual capability does not scale monotonically with parameter count.

Our findings establish physically grounded multimodal evaluation as a critical yet largely missing dimension of LVLM assessment. By bridging camera physics, human perceptual standards, and multimodal reasoning evaluation, SynDORBench enables systematic study of LVLM reliability under real-world perceptual constraints.

The main contributions of this work are summarized as follows:

\begin{enumerate}
    \item We introduce \textbf{SynDORBench}, the first DORI-calibrated benchmark for evaluating LVLM perceptual robustness under physically grounded imaging conditions. Unlike existing multimodal benchmarks that rely on dataset-relative visual difficulty, SynDORBench explicitly models camera distance, illumination, viewpoint, and pixel density according to measurable imaging physics and human perceptual thresholds.

    \item We propose a scalable \textbf{discernibility annotation framework} for low-visibility supervision under night and night-vision conditions. Using 2,000 human annotations, mask-conditioned perceptual features, and an RF+XGBoost ensemble, the framework propagates discernibility labels across a large synthetic corpus while preserving alignment with human perceptual judgments.

    \item We conduct a \textbf{large-scale robustness evaluation} across 18 state-of-the-art models, including open-source LVLMs, commercial LVLMs, and detector-based baselines. Our analysis reveals previously underexplored relationships between pixel density, physical imaging constraints, model scale, and multimodal perceptual reliability.

    \item We establish \textbf{physically grounded multimodal evaluation} as a new benchmark paradigm for studying LVLM robustness under real-world perceptual degradation, enabling direct comparison between algorithmic perception and human visibility capability thresholds.
\end{enumerate}

\section{Related Work}
\label{related_Work}

\paragraph{Synthetic Datasets} have become increasingly important for studying visual perception under controlled and reproducible conditions. Unlike real-world datasets, synthetic generation pipelines enable precise manipulation of scene geometry, illumination, camera pose, and environmental conditions, allowing researchers to isolate individual factors that influence model behavior. High-fidelity rendering frameworks such as Synscapes~\cite{wrenninge2018synscapes}, SPREAD~\cite{feng2025spread}, and SCaRL~\cite{ramesh2024scarl} demonstrate that synthetic imagery can reproduce physically plausible degradation effects arising from adverse weather, long-range observation, and changing illumination. Human-centric synthetic datasets such as SynBody~\cite{yang2023synbody} and Boundless~\cite{turkcan2024boundless} further show that synthetic data can effectively support downstream perception and human understanding tasks. Despite these advances, existing synthetic benchmarks primarily focus on semantic diversity, photorealism, or downstream task performance rather than physically grounded perceptual evaluation. In most prior work, visual degradation emerges implicitly from rendering configurations and is not explicitly tied to measurable imaging properties such as focal length, sensor geometry, aperture, viewing distance, or pixel density. Consequently, benchmark difficulty is typically dataset-relative rather than physically interpretable, limiting the ability to quantify how imaging constraints affect perceptual reliability.


\paragraph{Multimodal Benchmarking and Robustness Evaluation} such as MMBench~\cite{liu2024mmbench}, MMMU~\cite{yue2024mmmu}, SEED-Bench~\cite{li2023seed}, and MME~\cite{fu2025mme} have become standard evaluation suites for measuring multimodal reasoning, instruction following, and semantic understanding across diverse visual domains. While these benchmarks provide important insights into multimodal competence, they predominantly evaluate models under curated, high-visibility conditions where semantic information is largely preserved. As a result, current evaluations emphasize what LVLMs can recognize under idealized settings, while providing limited understanding of how perceptual degradation affects model reliability. Critical physical factors such as viewing distance, illumination variation, viewpoint distortion, and reduced pixel density remain largely underexplored despite their importance in real-world deployment scenarios. Several recent benchmarks partially address this limitation by introducing temporally grounded evaluation settings. Video-oriented datasets such as LVBench~\cite{wang2025lvbench} and CinePile~\cite{rawal2024cinepile} incorporate long-video reasoning, temporal dependencies, and action understanding. However, these datasets remain observational rather than physically controlled: degradation factors such as illumination, exposure, occlusion, and distance vary naturally but are not systematically parameterized according to measurable imaging constraints. Consequently, they can reveal aggregate model performance but cannot isolate the causal impact of specific physical degradation factors on multimodal perception. 


SynDORBench directly targets this missing dimension by introducing a physically calibrated benchmark for multimodal robustness evaluation. Rather than measuring semantic reasoning alone, SynDORBench systematically evaluates how LVLM perceptual capability degrades across DORI-calibrated visibility regimes under controlled variations in distance, lighting, and viewpoint. By connecting multimodal evaluation with camera physics and human perceptual standards, SynDORBench establishes a new benchmark paradigm for studying physically grounded perceptual robustness in LVLMs, as summarized in \cref{tab:bench_comparison}.

    \begin{table}[t]
    \small
    \centering
    \caption{Comparison of multimodal LVLM benchmarks.}
    \label{tab:bench_comparison}
    \scalebox{0.72}{
    \setlength{\tabcolsep}{5pt}
    \begin{tabular}{l p{2.2cm} p{2.8cm} c c c}
    \toprule
    \textbf{Benchmark} & \textbf{Modality} & \textbf{Answer Type} &
    \textbf{\# Models} & \textbf{\# QA Pairs} & \textbf{DORI-Calibrated Dist.} \\
    \midrule
    SEED-BENCH \cite{li2023seed}           & Image, Video & Multiple-choice & 18 & 19{,}242 & \xmark \\ 
    LV-BENCH  \cite{wang2025lvbench}           & Video & Multiple-choice & 27 & 1{,}549 & \xmark \\ 
    MMMU \cite{yue2024mmmu}               & Image & Multiple-choice, Open-ended & 24  & 11{,}550 & \xmark \\ 
    CinePile \cite{rawal2024cinepile}            & Video & Multiple-choice & 24 & 303{,}828 & \xmark \\ 
    LOKI \cite{ye2024loki}            & Video, Image, 3D, Audio & Multiple-choice, Open-ended & 28 & 18{,}000 & \xmark \\
    \rowcolor{gray!6}
    \textbf{SynDORBench-54k(Ours)} & Image & Multiple-choice, Open-ended &
    18 & 54,864 & \textbf{\cmark} \\
    \bottomrule
    \end{tabular}}
    \end{table}

\section{SynDORBench}
\label{SynDORBench}

    \begin{figure*}[t]
        \centering
        \includegraphics[width=0.92\linewidth]{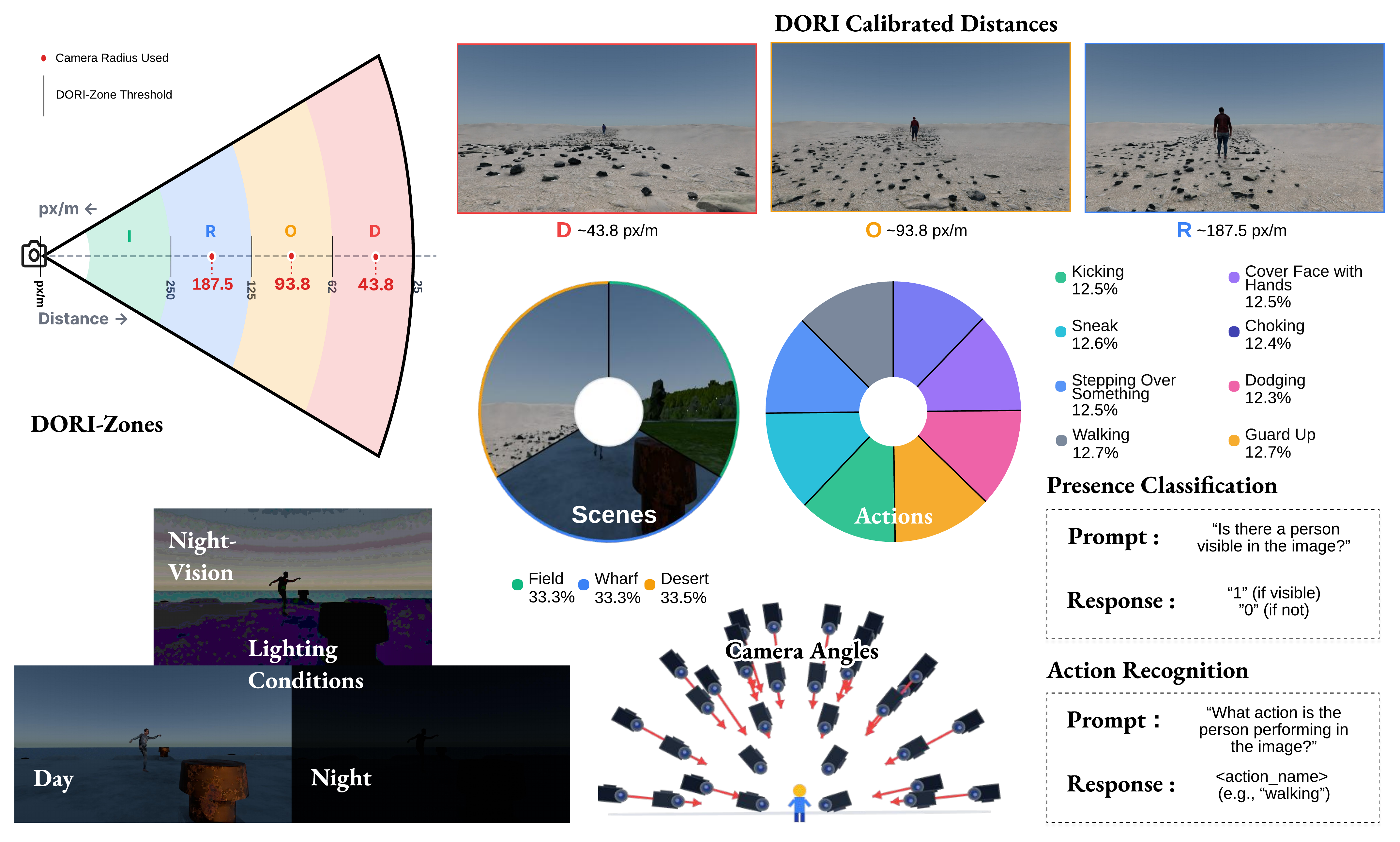}
        \vspace{-0.1in}
        \caption{Conceptual overview of SynDORBench: The benchmark systematically varies physical visibility factors (lighting, scene type, camera azimuth/elevation, action pose) and DORI-calibrated viewing distance, evaluating models on two tasks: 1) binary human-presence classification and 2) open-ended action recognition.}
        \label{fig:Concept banner}
    \end{figure*}
    
SynDORBench is a physically grounded evaluation framework designed to systematically study how LVLM perceptual capability degrades under progressively constrained imaging conditions. Unlike conventional multimodal benchmarks that primarily evaluate semantic understanding under curated visual settings, SynDORBench explicitly models the physical factors that govern visual evidence formation, including camera distance, illumination, viewpoint, and pixel density. By anchoring \underline{syn}thetic image generation to \underline{DO}RI-calibrated visibility \underline{r}ange thresholds, the \underline{bench}mark enables controlled analysis of multimodal perceptual robustness under measurable imaging constraints rather than uncontrolled dataset variation.

A central design principle of SynDORBench is \emph{factorized physical degradation}. Instead of relying on naturally occurring visual variability, we explicitly parameterize and isolate individual degradation factors to study their causal effect on semantic recoverability. This allows SynDORBench to characterize not only whether an LVLM succeeds or fails, but also under which physically interpretable visibility conditions perceptual breakdown occurs. The resulting framework bridges multimodal evaluation with camera physics and human perceptual standards, enabling direct comparison between algorithmic robustness and human visibility capability.

\subsection{Synthetic Image Creation}
\label{subsec:synth-img-creation}
This section describes the synthetic generation pipeline used to construct SynDORBench. We employ Blender as a controllable rendering environment to systematically manipulate scene geometry, camera configuration, illumination, viewpoint, and subject appearance while preserving physically interpretable imaging conditions.

\paragraph{Controlled Single-Human Design}

Although real-world image capturing environments often contain multiple individuals~\cite{chavdarova2018wildtrack}, SynDORBench intentionally adopts a single-human design to isolate the relationship between physical visibility constraints and perceptual capability. Human discernibility is fundamentally governed by the pixel density allocated to the entity of interest~\cite{bsen62676-4}. Introducing multiple subjects would entangle distance-dependent degradation with inter-person occlusion, crowd density, and spatial ambiguity, making it difficult to attribute performance degradation to specific imaging factors. Our controlled setup therefore enables causal analysis of how distance, lighting, and viewpoint independently affect LVLM perception under minimum-information regimes.

\paragraph{Physically Calibrated Camera Configuration}

The simulated camera uses a 1080p, 1/2.8" sensor, 2.8mm focal length, and f/1.4 aperture, reflecting common fixed-camera specifications. Full details on specification justification is provided in \cref{supsubsec:cam-spec}.



\paragraph{DORI-Calibrated Distance Modeling}

DORI-calibrated distances are derived from camera geometry and pixel-densities using the pinhole camera model~\cite{Hartley2004} and DORI perceptual framework~\cite{bsen62676-4} respectively. We sample midpoint pixel-density regimes between adjacent DORI zones: 43.8, 93.8, and 187.5~px/m, which converts to calibrated distances of 
23m, 10.8m, and 5.4m respectively. Full derivation is provided in \cref{supsubsec:model-distance}.

\paragraph{Camera Viewpoint Sampling}

To model realistic fixed-camera imaging viewing geometry, we adopt a hemispherical camera placement strategy parameterized by azimuth and elevation angles. The camera configuration consists of 8 azimuthal viewpoints and 4 elevation levels distributed around the subject. This design preserves consistent distance calibration while capturing viewpoint-dependent appearance variation.

\paragraph{Illumination Modeling}

SynDORBench incorporates four lighting conditions corresponding to morning, noon, evening, and midnight scenarios. The daytime settings vary solar azimuth and elevation angles to reproduce realistic directional illumination and shadow behavior, while the midnight configuration models extreme low-light visibility degradation. This allows controlled evaluation of LVLM robustness across progressively deteriorating illumination regimes.

\paragraph{Night-Vision Simulation}

Because camera systems frequently employ night-vision enhancement pipelines to improve visibility in low-light settings, SynDORBench additionally includes a night-vision subset derived from low-light renderings using a vendor-neutral image signal processing pipeline. More details found in \cref{supsubsec:night-vis-sim}.




\paragraph{Scene Diversity}

To reduce scene-specific bias while preserving physical controllability, SynDORBench includes three large-scale outdoor environments: a desert landscape, an open field, and a ship wharf. Assets are sourced from BlenderKit (RF/CC0) and selected to provide both environmental diversity and sufficient spatial extent for long-range DORI-calibrated rendering. The resulting environments vary substantially in texture density, background clutter, illumination interaction, and spatial composition.

\paragraph{Action Pose Generation}

Human action poses are derived from the BABEl dataset~\cite{BABEL:CVPR:2021}, built upon AMASS motion-capture corpus~\cite{AMASS}. Poses are filtered for abnormal activity perceptual relevance via a multi-stage embedding-based ranking and a final manual verification pipeline, then retargeted into Blender using SMPL-X~\cite{SMPL-X:2019}. Full details are provided in \cref{supsubsec:pose-filtering}.




\paragraph{Texture and Appearance Diversity}

To improve photometric diversity, we utilize the SMPLitex texture pack~\cite{casas2023smplitex}, which provides realistic variation in skin tone, clothing appearance, reflectance, and luminance characteristics. This diversity improves the robustness of subsequent perceptual evaluation by reducing overfitting to narrow appearance distributions and enabling richer light--surface interaction patterns.

\paragraph{Factor-Wise Stratified Sampling}


SynDORBench adopts factor-wise stratified sampling, where balancing is performed independently over each generative factor. For every factor category (e.g., DORI distance, lighting condition, texture), samples are drawn to maintain approximately uniform marginal distributions. This strategy provides broad coverage of the physical design space while avoiding severe over-representation of dominant factor combinations.

\subsection{Discernible Labels}
\label{subsec:discernible-labels}

    Although SynDORBench is synthetically generated, perceptual visibility is not binary. A human may be geometrically present within the image while remaining perceptually indiscernible due to insufficient pixel density, poor illumination, low contrast, or severe visibility degradation. To align presence classification labels with human perceptual capability, SynDORBench introduces discernibility labels that explicitly model whether a human annotator can reliably perceive the subject under rendered conditions. A pool of 2,000 was sampled using Cochran's formulation~\cite{cochran1977sampling} under a 95\% confidence interval and 5\% margin of error, of which 400 samples were held out as a test set. Enabled by the ground truth segmentation masks, we extracted mask-conditioned perceptual features capturing features that compared visual evidence inside and outside the mask, feature include: 1) unblocked area ratio, measuring the proportion of human pixels relative to the image. 2)Luminance Divergence, computed using the Kolmogorov--Smirnov (K-S) statistic~\cite{hodges1958significance} between foreground and background luminance distributions in the CIE Lab color space~\cite{iso_cie_11664_4_2019}. 3)Chromatic Distribution Distance, measured using Sliced Wasserstein distance~\cite{bonneel2015sliced} to quantify differences in hue and saturation distributions between the human region and the surrounding context. 4)Edge Discernibility, computed via the intersection-over-union between Sobel edge responses~\cite{sobel1968} inside the masked region and a dilated contour around the human boundary. To improve robustness under adverse illumination, each feature is computed for both the original rendering and its corresponding night-vision-enhanced variant. This allows the annotation model to jointly reason over raw luminance structure and enhanced visibility characteristics produced by the ISP pipeline. We formulate discernibility prediction as a binary perceptual classification problem. Hyperparameter optimization is performed using grid search with 5-fold cross-validation to improve generalization across visibility conditions. We evaluate both a decision-tree classifier and an ensemble model combining Random Forest and XGBoost. As shown in \cref{tab:annotation_model}, the ensemble model achieves the best overall performance and is subsequently used to propagate discernibility labels across the night and night-vision subsets of SynDORBench. Importantly, the discernibility framework allows SynDORBench to distinguish between \emph{physical presence} and \emph{perceptual visibility}, enabling more faithful evaluation of LVLM robustness under minimum-information imaging regimes.


    
    
    

    \begin{table*}[t]
    \centering
    \caption{Overview of evaluated models by scale and modality (Custom encoders denoted by * with limited publicly disclosed details).}
    \scalebox{0.7}{
    \begin{tabular}{l p{3.1cm} l l l l}
    \toprule
    \textbf{Model Type} & \textbf{Models} & \textbf{\# Params} & \textbf{Vision Encoder (VE)} & \textbf{VE Type} & \textbf{LLM Backbone} \\
    \midrule
    \textbf{Detection-only (non-LVLM)} & YOLO11x \cite{khanam2024yolov11} & 56.9M & -- & -- & -- \\
    \midrule
    \multirow{8}{*}{\textbf{Small LVLMs (SVLMs)} \cite{lu2024small}} 
     & SmolVLM2-256M-Video-Instruct \cite{marafioti2025smolvlm,smolvlm2_256m_video_instruct_2025} & 256M & SigLIP-B/16-512 & SigLIP & SmolLM2–135M \\
     & SmolVLM2-500M-Video-Instruct \cite{marafioti2025smolvlm, smolvlm2_500m_video_instruct_2025} & 500M & SigLIP-B/16-512 & SigLIP & SmolLM2–360M \\
     & InternVL3-1B-hf \cite{zhu2025internvl3} & 1B & InternViT-300M-448px-V2.5 & InternViT & Qwen2.5–0.5B \\
     & Perception-LM-1B \cite{cho2025perceptionlm}& 1B & PE L/14-448 & PE & Llama-3.2-1B \\
     & SmolVLM2-2.2B-Instruct \cite{marafioti2025smolvlm,smolvlm2_2.2b_instruct_2025} & 2.2B & SigLIP-So400M-patch14-384 & SigLIP & SmolLM2-1.7B \\
     & Perception-LM-3B \cite{cho2025perceptionlm} & 3B & PE L/14-448 & PE & Llama-3.2-3B \\
     & Qwen2.5-VL-3B-Instruct \cite{bai2025qwen2} & 3B & Custom ViT* & Custom ViT* & Qwen2.5-3B \\
     & gemma-3-4b-it \cite{team2025gemma} & 4B & SigLIP-So400M-896* & SigLIP & Gemma-3-4B \\
     & gemma-3n-E2B-it \cite{team2025gemma, google_gemma3n_devguide_2025} & 5B (2B) & MobileNet-V5-300M & MobileNet & Gemma-3n-E2B \\
    \midrule
    \multirow{7}{*}{\textbf{Medium LVLMs (MVLMs)}} 
     & Phi-4-multimodal-instruct \cite{abouelenin2025phi} & 6B & SigLIP-So400M* & SigLIP & Phi-4-mini \\
     & LLaVAction-7B \cite{ye2025llavaction} & 7B & SigLIP-so400m-patch14-384 & SigLIP & Qwen2-7B \\
     & LLaVA-NeXT-Video-7B-hf \cite{zhang2024llavanextvideo} & 7B & SigLIP-so400m-patch14-384 & SigLIP & Qwen2-7B \\
     & Molmo-7B-D-0924 \cite{deitke2025molmo} & 7B & CLIP-L/14 & CLIP & Qwen2-7B \\
     & Qwen2.5-VL-7B-Instruct \cite{bai2025qwen2} & 7B & Custom ViT* & Custom ViT* & Qwen2.5-7B \\
     & gemma-3n-E4B-it \cite{team2025gemma, google_gemma3n_devguide_2025} & 8B (4B) & MobileNet-V5-300M & MobileNet & Gemma-3n-E4B \\
     & InternVL3-8B-hf \cite{zhu2025internvl3} & 8B & InternViT-6B & InternViT & Qwen2.5-7B \\
    \midrule
    \textbf{Closed-source LVLMs} & gpt-4o-mini-2024-07-18 \cite{openai2024gpt4omini} & $\sim$8B \cite{abacha2024medec} & -- & -- & -- \\
    \bottomrule
    \end{tabular}
    }
    \label{tab:model_overview}
    \end{table*}

\subsection{Task Definition and Question Types}
\label{subsec:task-definition}
    Within this benchmark, a task is defined as a bounded perceptual objective that an LVLM must solve from a single image, under human capability-aligned imaging conditions. Each task is instantiated as a question template applied to every evaluated frame, every evaluated sample comprises one question and one expected answer. This yields a unified QA-based interface across heterogeneous LVLMs, allowing fair comparison irrespective of architectural differences, chat-template variations, or model-specific decoding protocols. The current benchmark focuses on two foundational perception tasks: 1) Human-Presence Classification (HPC), and 2) Action Recognition. These tasks were deliberately selected because they represent core long-range visual understanding capabilities commonly required in robotics, assistive technologies, and smart infrastructure. Further details of the action-recognition task and results will be found in the supplementary.
    

    For \textbf{HPC}, each LVLM receives the prompt: "Respond only with 1 or 0. Respond with 1 if a person is visible in the image, and 0 if not. Is there a person visible in the image?". This isolates grounded image-level binary HPC capability, and allows for evaluating sensitivity to distance, illumination, and occlusion effects without conflating perceptual capability with spatial localization. SynDORBench additionally includes mask and bounding-box metadata to support future human detection benchmarking.

\subsection{Evaluation Metrics}
\label{subsec:eval-metrics}

For the \textbf{HPC} task, we follow prior LVLM evaluation protocols that formulate existence queries as a binary classification problem~\cite{li2023evaluating,kaul2024throne}. We report both accuracy and macro-F1. Accuracy measures overall prediction correctness, while macro-F1 accounts for class imbalance and asymmetric error patterns, making it particularly informative for evaluating hallucination behavior under degraded visibility conditions~\cite{li2023evaluating}. Unlike conventional benchmarks that report aggregate performance alone, SynDORBench evaluates metrics across DORI-calibrated visibility regimes to characterize perceptual degradation under varying distance, illumination, and viewpoint conditions. To contextualize deployment reliability, we additionally define an \textbf{operational effectiveness threshold} of 80\% accuracy, motivated by prior real-world human detection literature~\cite{mokayed2023real,liu2012vehicle,velastin2020detecting,nguyen2021yolo}. Performance below this threshold is treated as operationally unreliable under the corresponding imaging conditions.

\section{Evaluation Results}
\label{evaluation_results}

In this section, we evaluate LVLM perceptual robustness on SynDORBench  
under progressively degraded DORI-calibrated imaging conditions. Additional stratified analyses are provided in \cref{apdx:HPC-lightings}.

\paragraph{Evaluated Models} We evaluate 16 open-source LVLMs, one commercial LVLM baseline, and YOLO11x, as summarized in \cref{tab:model_overview}. As a representative detection-only baseline, we adopt YOLO11x~\cite{khanam2024yolov11}, which remains widely used in detection systems~\cite{kukreti2025object}. YOLO11x is the strongest model in the YOLOv11 family and therefore provides a competitive detector-based reference point for comparison against multimodal architectures. We additionally include GPT-4o-mini~\cite{openai2024gpt4omini} as a closed-source commercial LVLM baseline. Because SynDORBench focuses on perceptual robustness rather than large-scale reasoning capacity, we restrict evaluation to models with $\leq$8B parameters to maintain practical relevance for low-latency visual understanding systems. Larger models incur substantially higher inference costs, making direct comparison with real-time detection pipelines less meaningful~\cite{ultralytics2024benchmark}. The evaluated open-source models include SmolVLM2~\cite{marafioti2025smolvlm}, InternVL3~\cite{zhu2025internvl3}, Perception-LM~\cite{cho2025perceptionlm}, Qwen2.5-VL~\cite{bai2025qwen2}, Gemma-3~\cite{team2025gemma}, Phi-4-multimodal~\cite{abouelenin2025phi}, LLaVAction~\cite{ye2025llavaction}, LLaVA-NeXT-Video~\cite{zhang2024llavanextvideo}, and Molmo~\cite{deitke2025molmo}. Following the categorization proposed by Lu et al.~\cite{lu2024small}, we group models with $\leq5$B parameters as Small Vision--Language Models (SVLMs), while models in the 6--8B range are grouped as Medium Vision--Language Models (MVLMs).

\paragraph{HPC Robustness Across Distances} To isolate the effect of perceptual distance degradation, we evaluate all models across DORI-calibrated distance regimes while averaging over other generative factors such as illumination and viewpoint. We additionally report group means for SVLMs and MVLMs to analyze scale-dependent robustness trends. Results are summarized in \cref{tab:over_distance_full}.

Across all evaluated models, HPC performance exhibits clear distance sensitivity: both accuracy and macro-F1 consistently decrease as viewing distance increases from 5.4\,m to 23.0\,m. This confirms that reduced pixel density substantially constrains semantic recoverability under physically degraded imaging conditions.

MVLMs achieve the strongest overall group performance across all distances, indicating that increased model capacity generally improves perceptual robustness. However, the relationship between parameter count and robustness is not strictly monotonic. Several compact SVLMs, particularly \textit{gemma-3-4b-it} and \textit{gemma-3n-E2B-it}, match or exceed the performance of larger MVLMs, suggesting that architectural design and multimodal alignment may play a more important role than scale alone under low-information regimes.

Compared to LVLMs, YOLO11x exhibits substantially weaker robustness, particularly at longer viewing distances. Importantly, this degradation is not driven solely by distance. YOLO11x already falls below the 80\% operational reliability threshold at 5.4\,m, where visibility conditions remain relatively favorable within SynDORBench. This suggests that conventional detector-based pipelines are highly sensitive not only to pixel-density reduction, but also to heterogeneous illumination, contrast variation, and perceptual ambiguity.

Overall, the results indicate that modern LVLMs substantially outperform traditional detector-based systems on DORI-calibrated HPC and remain competitive with commercial closed-source LVLMs under progressively degraded visibility conditions.

    \begin{table}[t]
    \caption{Performance degradation across DORI-calibrated distances for presence classification, with accuracy (Acc) and macro-F1 (F1) reported. Best scores are in bold.}
    \label{tab:over_distance_full}
    \centering
    \scalebox{0.81}{
    \begin{tabular}{p{5cm}cccccc}
    \toprule
    \textbf{Model} & \multicolumn{2}{c}{\textbf{5.4 m}} & \multicolumn{2}{c}{\textbf{10.8m} } & \multicolumn{2}{c}{\textbf{23.0 m}} \\
    \cmidrule(lr){2-3} \cmidrule(lr){4-5} \cmidrule(lr){6-7}
     & \textbf{Acc} & \textbf{F1} & \textbf{Acc} & \textbf{F1} & \textbf{Acc} & \textbf{F1} \\
    \midrule
    YOLO11x & 73.92 & 50.17 & 65.38 & 50.36 & 50.29 & 42.98 \\
    SmolVLM2-256M-Video-
Instruct  & 10.02 & 9.63 & 13.47 & 12.68 & 22.70 & 18.69 \\
    SmolVLM2-500M-Video-
Instruct & 23.20 & 20.23 & 26.51 & 24.04 & 27.19 & 22.08 \\
    InternVL3-1B-hf & 87.77 & 59.63 & 85.69 & 64.58 & 83.97 & 68.13 \\
    Perception-LM-1B & 69.77 & 48.21 & 62.64 & 49.11 & 45.94 & 39.89 \\
    SmolVLM2-2.2B-Instruct & 80.84 & 55.07 & 75.89 & 57.64 & 66.51 & 56.22 \\
    Perception-LM-3B & 75.76 & 50.83 & 71.76 & 53.77 & 58.80 & 48.37 \\
    Qwen2.5-VL-3B-Instruct & 66.70 & 45.49 & 64.00 & 48.86 & 54.75 & 45.16 \\
    gemma-3-4b-it  & \textbf{94.66} & 50.44 & \textbf{92.91} & 54.04 & 85.66 & 67.68 \\
    gemma-3n-E2B-it  & 87.10 & \textbf{75.78} & 86.29 & \textbf{65.27} & \textbf{88.87} & \textbf{72.12} \\
    \midrule
    \textbf{SVLM (mean)} & 66.20 & 46.14 & 64.35 & 47.78 & 59.38 & 48.71 \\
    \midrule
    Phi-4-multimodal-instruct & 84.97 & 56.59 & 83.42 & 62.74 & 79.53 & 65.76 \\
    LLaVAction-7B & 81.80 & 55.96 & 74.70 & 57.33 & 57.88 & 50.01 \\
    LLaVA-NeXT-Video-7B-
hf  & 81.99 & 54.84 & 73.15 & 54.76 & 54.90 & 47.17 \\
    Molmo-7B-D-0924 & 84.63 & 57.68 & 80.61 & 60.82 & 70.76 & 59.06 \\
    Qwen2.5-VL-7B-Instruct & 86.13 & 58.41 & 84.75 & 63.86 & 80.24 & 64.99 \\
    gemma-3n-E4B-it  & 87.77 & 59.08 & 86.73 & 65.05 & 87.76 & 70.40 \\
    InternVL3-8B-hf  & 85.45 & 58.08 & 83.13 & 62.67 & 71.66 & 59.84 \\
    \midrule
    \textbf{MVLM (mean)} & 84.68 & 57.24 & 80.93 & 61.03 & 71.82 & 59.61 \\
    \midrule
    gpt-4o-mini-2024-07-18 & 82.72 & 55.79 & 77.87 & 58.72 & 69.60 & 59.03 \\
    \bottomrule
    \end{tabular}}
    \end{table}
    
\paragraph{Discussion}

Interestingly, while MVLMs achieve the strongest overall group performance, robustness does not scale monotonically with parameter count. Several compact SVLMs remain highly competitive and, in some cases, outperform larger models under degraded visibility conditions. This suggests that multimodal robustness depends not only on model scale, but also on factors such as vision encoder design, visual alignment quality, and robustness of multimodal feature integration. These findings indicate that physically grounded perception may require architectural properties distinct from those optimized by conventional semantic reasoning benchmarks.

Another notable observation is the substantial robustness gap between LVLMs and traditional detector-based systems. YOLO11x exhibits rapid degradation under long-range and heterogeneous illumination conditions, even falling below operational reliability thresholds at relatively favorable visibility regimes. In contrast, LVLMs retain stronger discriminability under the same conditions, potentially because multimodal architectures can leverage broader contextual and semantic priors rather than relying solely on local appearance cues. This suggests that modern LVLMs may offer a more resilient perception framework for low-information environments where conventional detectors become brittle.

\section{Conclusion}
\label{Discussion}

In this paper, we introduced SynDORBench, the first physically grounded benchmark for evaluating LVLM perceptual robustness under DORI-calibrated imaging conditions. Unlike existing multimodal benchmarks that primarily assess semantic reasoning under ideal visual settings, SynDORBench systematically studies how multimodal perception degrades as visual evidence deteriorates under controlled physical constraints.

Through large-scale evaluation across open-source LVLMs, commercial LVLMs, and detector-based baselines, we show that pixel density and imaging conditions play a central role in determining multimodal perceptual reliability. Our experiments further demonstrate that modern LVLMs substantially outperform traditional detector-based pipelines under long-range and low-visibility conditions, while several compact open-source models achieve robustness comparable to larger commercial systems.

Overall, SynDORBench establishes physically grounded multimodal evaluation as an important and previously underexplored dimension of LVLM assessment. We hope the benchmark will support future research on robust multimodal perception, physically constrained reasoning, and reliability-aware evaluation for real-world vision-language systems.

\textbf{Data Availability}
        The dataset and evaluation framework will be released under the CC BY-NC-SA 4.0 license for non-commercial academic research. The release includes rendered images, metadata, and evaluation code to support reproducibility and future benchmarking research, while respecting the licensing terms of upstream resources including SMPL-X, AMASS, BABEL, and SMPLitex.



\textbf{Limitations} SynDORBench is designed as a controlled and physically grounded benchmark for studying how imaging conditions affect LVLM perceptual robustness. To enable reproducible and factorized analysis, the current benchmark focuses on single-human, frame-level scenarios under systematically calibrated visibility conditions. 




\textbf{Broader Impact}
SynDORBench is intended to advance research on reliable multimodal perception under adverse imaging conditions, with potential applications in robotics, assistive systems, intelligent infrastructure, autonomous platforms, and safety-critical visual understanding. 


\textbf{Safeguards}  
    SynDORBench consists entirely of synthetically rendered imagery and does not contain real individuals or real-world imagery, thereby avoiding privacy and identity-related concerns associated with real data collection. The benchmark is released strictly for non-commercial academic research purposes to encourage responsible use within the research community.
    

\newpage

\section*{Acknowledgments}
This work is supported by the SIT-NVIDIA AI Centre.

\bibliographystyle{splncs04}
\bibliography{main}





\newpage
\appendix

\section{Technical Appendices and Supplementary Material}



\subsection{Additional Synthetic Image Creation Details}

\subsubsection{Camera Specification Survey}
\label{supsubsec:cam-spec}

    The camera specification survey in \cref{tab:camera_specifications} details the typical configuration in fixed-camera systems. Across major manufacturers(Axis \cite{Axis2025ProductComparisonQ2}, Hikvision \cite{Hikvision2023PQG}, Dahua \cite{Dahua2020ProductCatalog}), the dominant hardware profile consists of 1/2.8" sensors paired with fixed 2.8mm lenses. This configuration is used for our simulated hardware to capture the images in Blender. 

    \begin{table*}[h]
    \centering
    \caption{Specifications of representative commercial camera models commonly used in fixed-camera systems.}
    \label{tab:camera_specifications}
    \resizebox{\linewidth}{!}{
    \begin{tabular}{l l c c c c c}
    \toprule
    \textbf{Manufacturer} & \textbf{Model} & \textbf{Resolution} & \textbf{Sensor} & 
    \textbf{Lens Type} & \textbf{Focal Length (mm)} & \textbf{Aperture} \\
    \midrule

    Axis & Q1615-LE Mk III & 1920×1080 & 1/2.8'' & Varifocal & 2.8--8.5   & f/1.2 \\
    Axis & Q1715             & 1920×1080 & 1/2.8'' & Varifocal & 4--84.6    & f/1.6--4.5 \\
    Axis & P1385             & 1920×1080 & 1/2.8'' & Varifocal & 2.8--13    & f/1.4 \\
    Axis & P1385-B           & 1920×1080 & 1/2.8'' & --         & --         & -- \\
    Axis & P1385-E           & 1920×1080 & 1/2.8'' & Varifocal & 2.8--13    & f/1.4 \\
    Axis & P1385-BE          & 1920×1080 & 1/2.8'' & --         & --         & -- \\
    
    \midrule
    
    Hikvision & HX0D12Z2S    & 1920×1080 & 1/2.8'' & Varifocal & 2.7--13.5  & f/1.4 \\
    Hikvision & HX-B12Z2S    & 1920×1080 & 1/2.8'' & Varifocal & 2.7--13.5  & f/1.4 \\
    Hikvision & HX-D12F2S    & 1920×1080 & 1/2.8'' & Fixed     & 2.8        & f/1.4 \\
    Hikvision & HX-T12F6S    & 1920×1080 & 1/2.8'' & Fixed     & 2.8        & f/1.4 \\
    Hikvision & HX-B12F2S    & 1920×1080 & 1/2.8'' & Fixed     & 2.8        & f/1.4 \\
    Hikvision & HX-LB12F6S   & 1920×1080 & 1/2.8'' & Fixed     & 2.8        & f/1.4 \\
    
    \midrule
    
    DaHua & N25CL5Z          & 1920×1080 & 1/2.8'' & Varifocal & 2.7--13.5  & f/1.4 \\
    DaHua & N24CL52          & 1920×1080 & 1/2.8'' & Fixed     & 2.8        & f/1.6 \\
    DaHua & DH-IPC-HDBW4239RN-ASE & 1920×1080 & 1/2.8'' & Fixed & 3.6        & f/1.0 \\
    DaHua & N24BN52          & 1920×1080 & 1/2.8'' & Fixed     & 2.8        & f/2.0 \\
    DaHua & N22AL12          & 1920×1080 & 1/2.7'' & Fixed     & 2.8        & f/2.0 \\
    DaHua & N28BL7Z          & 1920×1080 & 1/1.9'' & Varifocal & 4.1--16.4  & f/1.53 \\
    
    \bottomrule
    \end{tabular}}
    \end{table*}

\subsubsection{Action Pose Filtering Pipeline}
\label{supsubsec:pose-filtering}

    To identify relevant visually discernible actions from the BABEL dataset~\cite{BABEL:CVPR:2021}, a three-stage filtering process was applied. First, to reduce the excessive number of frames, frame-level BABEL action descriptions were randomly shuffled, and 100 frames were sampled to get a diverse representation of frame labeled actions. The sampled frames were embedded using E5-large-v2\cite{wang2022text}, and cosine similarity was computed against the phrase "outdoor abnormal action" to obtain an initial relevance ranking. Candidate actions are subsequently reranked by Gemma-3-13b-it~\cite{team2025gemma} according to action relevance. Finally a manual verification step is performed to remove semantically inconsistent labels and to align the labels with human perception from a still frame. 

    The resulting subset prioritizes visually distinguishable static actions suitable for single-frame perception evaluation. The selected poses are retargeted and animated within Blender. The final action set comprises: walking, kicking, sneaking, choking, dodging and covering face with hands, stepping over something and guard up.

\subsubsection{Discernible Label Model Ablation}
    \begin{table}[h]
    \small
    \centering
    \caption{Performance of discernibility annotation models on validation set (5-Fold cross-validation).}
    \label{tab:annotation_model}
    \scalebox{0.82}{
    \begin{tabular}{lcc}
    \toprule
    \textbf{Model} & \textbf{Macro F1 (\%)} & \textbf{Accuracy (\%)} \\
    \midrule
    Decision Tree & 82.41 & 90.25 \\
    RF + XGBoost Ensemble & \textbf{83.55} & \textbf{90.50} \\
    \bottomrule
    \end{tabular}}
    \end{table}
    
\subsubsection{Modeling Distance by Pixel Density}
\label{supsubsec:model-distance}
Under the pinhole camera model~\cite{Hartley2004}, the vertical field of view (VFOV) is defined as:

\begin{equation}
\theta_{\mathrm{VFOV}} =
2 \tan^{-1}\!\left(
\frac{h_{\mathrm{sensor}}}{2f}
\right),
\label{eq:vfov_angle}
\end{equation}

where $h_{\mathrm{sensor}}$ denotes the sensor height and $f$ the focal length. At distance $D$, the corresponding vertical scene coverage is:

\begin{equation}
h_{\mathrm{scene}} =
2D \tan\!\left(
\frac{\theta_{\mathrm{VFOV}}}{2}
\right),
\label{eq:vfov_height}
\end{equation}

and the resulting vertical pixel density is given by:

\begin{equation}
PPM_{\mathrm{vert}} =
\frac{h_{\mathrm{res}}}{h_{\mathrm{scene}}},
\label{eq:ppm_vert}
\end{equation}

where $h_{\mathrm{res}}$ denotes the image height in pixels. Rearranging yields the viewing distance as a function of the required pixel density:

\begin{equation}
D =
\frac{h_{\mathrm{res}}}
{PPM_{\mathrm{req}}
\cdot
2 \tan\!\left(
\frac{\theta_{\mathrm{VFOV}}}{2}
\right)},
\label{eq:dori_distance}
\end{equation}

where $PPM_{\mathrm{req}}$ denotes the target DORI pixel density threshold.

\subsubsection{Night-Vision Simulation}
\label{supsubsec:night-vis-sim}

Real commercial night-vision systems rely on proprietary image signal processing (ISP) pipelines that are typically inaccessible~\cite{bielova2019digital}. To approximate these systems, we implement a vendor-neutral ISP pipeline inspired by prior computational imaging work~\cite{bielova2019digital,heide2014flexisp}.

The pipeline consists of:
(1) sRGB linearization,
(2) automatic gain control (AGC) normalization~\cite{fowler2004automatic},
(3) color normalization~\cite{buchsbaum1980spatial},
(4) BM3D denoising~\cite{dabov2007image,makinen2020collaborative},
(5) local contrast enhancement using CLAHE~\cite{mishra2021contrast}, and
(6) tone mapping~\cite{reinhard_tone_mapping}.

This process produces physically plausible night-vision imagery while preserving consistency with the underlying DORI-calibrated visibility conditions.

\subsection{Additional Task Definition and Question Type}
\label{apdx:act-reg-task}
\paragraph{Action Recognition} aims to determine if a model is able to identify the specific activity performed by the human subject presented in each image. This task evaluates an LVLM’s ability to interpret human pose, motion cues, and contextual elements necessary for recognizing semantically distinct behaviors. Each image is paired with an instruction and open-ended question (“Respond only with the name of the action. What action is the person performing in the image?”) requiring a short natural-language response such as walking, running, or waving. Unlike the binary HPC task, action recognition emphasizes higher-level visual understanding, testing a model’s ability to associate human posture and spatial configuration with the corresponding action category. It also provides insight into the model’s robustness under DORI-calibrated conditions, where reduced visibility, distance, and illumination can degrade action-level semantic recognition.

\subsection{Additional Metrics}

For the \textbf{Action Recognition} task, we follow prior LVLM evaluation protocols that treat open-ended responses as text predictions and therefore evaluate semantic agreement beyond exact string matching \cite{hessel2021clipscore}. For open-ended action recognition, our prompts explicitly instruct the model to respond concisely with the action, resulting in outputs that are typically 1-3 tokens. In this regime, lexical-overlap metrics (e.g., ROUGE-L) become unstable. Since utility depends on semantic proximity rather than exact lexical reproduction, we evaluate using E5Score \cite{wang2022text} to quantify sentence-level semantic alignment, following the same embedding-similarity paradigm as CLIPScore \cite{hessel2021clipscore} and SigLIPScore which is commonly used for Image-Text alignment \cite{neau2025measuring}, but modifying to fit our Text-to-Text setting. We use E5Score to avoid score bias due to the LVLMs using the vision encoders as the scoring metrics (e.g., CLIP, SigLIP). To improve interpretability, we report a distance-based z-score that highlights relative performance differences across models. Additionally, we obtain a human baseline from one non-expert annotator on a uniformly random sample of 300 instances.

\subsection{Computational Resources}
\label{computational-resources}
Raw image rendering was conducted on 4 NVIDIA A6000 GPUs, totaling approximately 535.7 GPU hours. Night-vision enhanced image creation was conducted on the CPU, totaling approximately 19 CPU hours. Experimental evaluations were conducted on 2--3 NVIDIA L40S GPUs, depending on availability. Each model required approximately 4 GPU hours across both tasks, totaling approximately 72 GPU hours for the full evaluation across 18 models.

\subsection{Additional Results}

\subsubsection{Action Recognition Robustness Across Distances}
\label{apdx:act-reg-distances}

    \begin{table}[h]
    \caption{Action recognition performance across DORI distances. Mean E5Score and corresponding z-score per distance. Best z-scores are in bold.}
    \label{tab:actreg_over_distance_full}
    \vspace{+0.1in}
    \centering
    \scalebox{0.81}{
    \begin{tabular}{p{5cm}ccc}
    \toprule
    \textbf{Model} & \textbf{5.4 m} & \textbf{10.8 m} & \textbf{23.0 m} \\
    \midrule
    Non-Expert Human & 0.94 & 0.91 & 0.83 \\
    \midrule
    SmolVLM2-256M-Video-Instruct & 0.78 (-1.45) & 0.77 (-1.39) & 0.77 (-1.15) \\
    SmolVLM2-500M-Video-Instruct  & 0.80 (-0.76) & 0.79 (-0.75) & 0.78 (-0.69) \\
    Internvl3-1B-hf & 0.81 (-0.23) & 0.80 (-0.43) & 0.78 (-0.72) \\
    Perception-LM-1B & 0.77 (-1.87) & 0.76 (-1.78) & 0.75 (-1.72) \\
    SmolVLM2-2.2B-Instruct & 0.84 (0.93) & 0.83 (0.98) & 0.83 (1.09) \\
    Perception-LM-3B  & 0.77 (-1.98) & 0.76 (-1.98) & 0.75 (-1.83) \\
    Qwen2.5-VL-3B-Instruct  & 0.83 (0.69) & 0.82 (0.61) & 0.82 (0.65) \\
    gemma-3-4b-it  & 0.80 (-0.59) & 0.80 (-0.52) & 0.78 (-0.63) \\
    gemma-3n-E2B-it & 0.84 (\textbf{1.04}) & 0.84 (\textbf{1.14}) & 0.83 (\textbf{1.13}) \\
    \midrule
    \textbf{SVLM (mean)} & 0.80 (-0.79) & 0.80 (-0.65) & 0.79 (-0.39) \\
    \midrule
    Phi4-multimodal-instruct & 0.84 (0.90) & 0.83 (0.95) & 0.83 (1.12) \\
    LLaVAction-7B & 0.84 (0.89) & 0.83 (0.95) & 0.83 (0.98) \\
    LLaVA-NeXT-Video-7B-hf  & 0.82 (-0.02) & 0.80 (-0.23) & 0.78 (-0.64) \\
    Molmo-7B-D-0924 & 0.82 (0.23) & 0.82 (0.39) & 0.82 (0.51) \\
    Qwen2.5-VL-7B-Instruct & 0.84 (0.88) & 0.83 (0.88) & 0.83 (0.90) \\
    gemma-3n-E4B-it  & 0.83 (0.57) & 0.82 (0.64) & 0.82 (0.59) \\
    internVL3-8B-hf  & 0.82 (0.34) & 0.81 (0.17) & 0.81 (0.18) \\
    \midrule
    \textbf{MVLM (mean)} & 0.83 (0.44) & 0.82 (0.59) & 0.82 (0.53) \\
    \midrule
    gpt-4o-mini-2024-07-18 & 0.83 (0.44) & 0.82 (0.39) & 0.81 (0.23) \\
    \bottomrule
    \end{tabular}}
    \end{table}

    We evaluate all LVLMs on SynDORBench-54k and then average over non-distance factor similarly to before, so that the reported scores reflect the effect of distance alone. Results are detailed in \cref{tab:actreg_over_distance_full}. Across all three distances, we observe a monotonic decrease in performance as radius increases, indicating that action recognition is strongly distance-sensitive and degrades as pixel density falls. For group means, the MLVLMs exceed that of the SVLMs at all radii, suggesting that parameter count continues to provide a net advantage for open-ended action classification when visual detail becomes sparse. However, this effect is not purely size-deterministic, although MLVLMs have higher group means, Gemma-3n-E2B-it, despite not being the largest model, achieves the best individual performance across all three distances. This indicates that architectural and alignment choices can yield better semantic retention than scale alone. Notably, the mean MLVLM exhibits a similar level of performance to GPT-4o mini, indicating that open medium-scale architectures can approach commercial-grade action recognition at the DORI ranges considered. However, it should be noted that all LVLMs fail to surpass a non-expert human on all radii, highlighting a clear gap in capability on this task.

\subsubsection{HPC Performance Over Lightings}
\label{apdx:HPC-lightings}
    To complement the main-manuscript analysis, we conduct a series of additional evaluations that probe how model performance varies across distance, lighting, and pixel-density regimes, offering a deeper view of robustness characteristics not captured in aggregate metrics.
    
    \begin{landscape}
    \begin{center}
    \captionof{table}{HPC performance across DORI-calibrated distances (5.4m, 10.8m, 23.0m). We report accuracy and F1-score for all models. The best scores for each column are in bold; underlined numbers indicate the second-highest.}
    \label{tab:hpc distance lighting}
    \footnotesize
    \centering
    \resizebox{\linewidth}{!}{%
    {\scriptsize
    \begin{tabular}{
        p{2cm}p{0.6cm}p{0.6cm}p{0.6cm}p{0.6cm}|
        p{0.6cm}p{0.6cm}p{0.6cm}p{0.6cm}p{0.6cm}p{0.6cm}p{0.6cm}p{0.6cm}|
        p{0.6cm}p{0.6cm}p{0.6cm}p{0.6cm}p{0.6cm}p{0.6cm}p{0.6cm}p{0.6cm}}
    \hline
                                 & \multicolumn{4}{c|}{\textbf{day}}        & \multicolumn{8}{c|}{\textbf{night}}                                                                             & \multicolumn{8}{c}{\textbf{night-vision}}                                                                      \\ \hline
    \textbf{radius(m)}                    & \textbf{5.4}    & \textbf{10.8}  & \textbf{23}     & \textbf{avg}   & \multicolumn{2}{c}{\textbf{5.4}} & \multicolumn{2}{c}{\textbf{10.8}} & \multicolumn{2}{c}{\textbf{23}} & \multicolumn{2}{c|}{\textbf{avg}} & \multicolumn{2}{c}{\textbf{5.4}} & \multicolumn{2}{c}{\textbf{10.8}} & \multicolumn{2}{c}{\textbf{23}} & \multicolumn{2}{c}{\textbf{avg}} \\ \hline
    \textbf{metrics}                      & \textbf{acc}    & \textbf{acc}   & \textbf{acc}    & \textbf{acc}   & \textbf{acc}       & \textbf{f1-score}    & \textbf{acc}       & \textbf{f1-score}     & \textbf{acc}      & \textbf{f1-score}    & \textbf{acc}       & \textbf{f1-score}     & \textbf{acc}       & \textbf{f1-score}    & \textbf{acc}       & \textbf{f1-score}     & \textbf{acc}      & \textbf{f1-score}    & \textbf{acc}       & \textbf{f1-score}    \\ \hline
    YOLO11x                      & 95.06  & 85.42 & 53.49  & 77.99 & 62.80     & 50.51       & 54.29     & 51.65        & 47.00    & 44.96       & 54.70     & 49.04        & 63.90     & 51.25       & 56.43     & 53.36        & 50.39    & 49.11       & 56.91     & 51.24       \\
    SmolVLM2-256M-Video-Instruct & 9.88   & 8.54  & 7.79   & 8.74  & 12.48     & 12.44       & 15.20     & 15.14        & 27.02    & 23.47       & 18.23     & 17.02        & 7.71      & 7.46        & 16.68     & 15.03        & 33.31    & 25.37       & 19.23     & 15.95       \\
    SmolVLM2-500M-Video-Instruct & 23.49  & 26.40 & 11.53  & 20.47 & 7.82      & 7.57        & 16.87     & 15.20        & 33.11    & 24.88       & 19.27     & 15.88        & 38.29     & 34.09       & 36.24     & 36.05        & 36.93    & 31.02       & 37.16     & 33.72       \\
    Internvl3-1B-hf                 & 99.95  & 99.77 & 95.09  & 98.27 & 78.20     & 61.48       & 74.94     & 68.32        & 71.96    & 71.49       & 75.03     & 67.10        & 85.17     & 67.41       & 82.35     & \underline{75.46}        & \underline{84.87}    & \underline{84.16}       & 84.13     & \underline{75.68}       \\
    Perception-LM-1B             & 90.80  & 78.79 & 40.33  & 69.97 & 43.88     & 37.99       & 39.05     & 38.65        & 35.92    & 29.44       & 39.62     & 35.36        & 74.64     & 59.07       & 70.07     & 64.61        & 61.58    & 61.50       & 68.76     & 61.73       \\
    SmolVLM2-2.2B-Instruct                & 95.73  & 92.02 & 71.57  & 86.44 & 64.79     & 51.92       & 57.95     & 54.62        & 54.37    & 53.72       & 59.04     & 53.42        & 82.01     & 64.39       & 77.71     & 70.39        & 73.60    & 73.24       & 77.77     & 69.34       \\
    Perception-LM-3B             & 99.33  & 95.42 & 70.20  & 88.31 & 60.82     & 49.23       & 52.96     & 50.57        & 46.53    & 44.37       & 53.44     & 48.06        & 67.14     & 53.43       & 66.91     & 61.92        & 59.67    & 59.50       & 64.58     & 58.29       \\
    Qwen2.5-VL-3B-Instruct                & 94.10  & 86.68 & 63.12  & 81.30 & 39.42     & 34.89       & 39.56     & 39.10        & 40.44    & 36.20       & 39.81     & 36.73        & 66.57     & 53.09       & 65.78     & 61.03        & 60.69    & 60.57       & 64.35     & 58.23       \\
    gemma-3-4b-it                    & \underline{99.98}  & \textbf{99.95} & \textbf{100.00} & \textbf{99.97} & \textbf{92.67}     & 51.26       & \textbf{90.71}     & 57.49        & 75.59    & 48.54       & \textbf{86.32}     & 52.43        & \textbf{91.33}     & 50.05       & \textbf{88.07}     & 54.64        & 81.40    & 54.49       & \textbf{86.93}     & 53.06       \\
    gemma-3n-E2B-it                  & \textbf{100.00} & \underline{99.72} & 98.75  & 99.49 & 75.08     & 58.11       & 76.42     & 70.11        & \underline{79.49}    & \underline{79.08}       & 77.00     & 69.10        & 86.22     & \textbf{69.23}       & \underline{82.74}     & \textbf{75.75}        & \textbf{88.38}    & \textbf{87.61}       & \underline{85.78}     & \textbf{77.53}       \\
    phi4-multimodal-instruct     & 99.93  & 98.51 & 86.10  & 94.85 & 72.85     & 57.54       & 71.59     & 65.89        & 75.70    & 75.44       & 73.38     & 66.29        & 82.13     & 62.25       & 80.16     & 72.71        & 76.79    & 75.57       & 79.70     & 70.18       \\
    LLaVAction-7B                   & 97.20  & 89.92 & 58.74  & 81.95 & 65.07     & 52.11       & 56.78     & 53.67        & 47.70    & 45.84       & 56.52     & 50.54        & 83.14     & 66.49       & 77.40     & 70.99        & 67.20    & 67.19       & 75.91     & 68.22       \\
    LLaVA-NeXT-Video-7B-hf           & 96.69  & 87.48 & 56.86  & 80.35 & 68.88     & 54.73       & 59.94     & 56.22        & 52.77    & 51.95       & 60.53     & 54.30        & 80.39     & 60.65       & 72.02     & 61.40        & 55.07    & 53.31       & 69.16     & 58.45       \\
    Molmo-7B-D-0924              & 99.81  & 97.36 & 78.82  & 92.00 & 69.85     & 55.49       & 66.56     & 61.68        & 59.48    & 59.29       & 65.30     & 58.82        & 84.24     & 67.58       & 77.90     & 71.46        & 73.99    & 73.82       & 78.71     & 70.96       \\
    Qwen2.5-VL-7B-Instruct                & 99.93  & 99.57 & 93.80  & 97.77 & \underline{81.24}     & \textbf{64.19}       & \underline{79.15}     & \textbf{72.43}        & 75.66    & 75.42       & 78.68     & \underline{70.68}        & 77.23     & 61.07       & 75.53     & 69.27        & 71.26    & 71.14       & 74.67     & 67.16       \\
    gemma-3n-E4B-it                  & \underline{99.98}  & 99.55 & \underline{99.07}  & \underline{99.53} & 76.46     & 59.28       & 78.29     & \underline{71.80}        & \textbf{81.71}    & \textbf{81.18}       & \underline{78.82}     & \textbf{70.75}        & \underline{86.87}     & \underline{67.97}       & 82.35     & 73.46        & 82.49    & 80.25       & 83.90     & 73.89       \\
    internVL3-8B-hf                 & \underline{99.98}  & 98.97 & 79.58  & 92.84 & 72.49     & 57.46       & 70.42     & 64.91        & 62.32    & 62.27       & 68.41     & 61.54        & 83.87     & 66.79       & 80.01     & 73.37        & 73.09    & 72.95       & 78.99     & 71.04       \\
    gpt-4o-mini-2024-07-18       & 98.97  & 95.82 & 73.84  & 89.54 & 69.25     & 55.05       & 63.21     & 58.91        & 60.34    & 60.20       & 64.26     & 58.05        & 79.94     & 62.59       & 74.59     & 68.31        & 74.61    & 74.42       & 76.38     & 68.44  
    \\
    \hline
    \end{tabular}
    }
    }
    \end{center}
    \end{landscape}

    \begin{landscape}
        \begin{figure}
        \centering
        \includegraphics[width=1\linewidth]{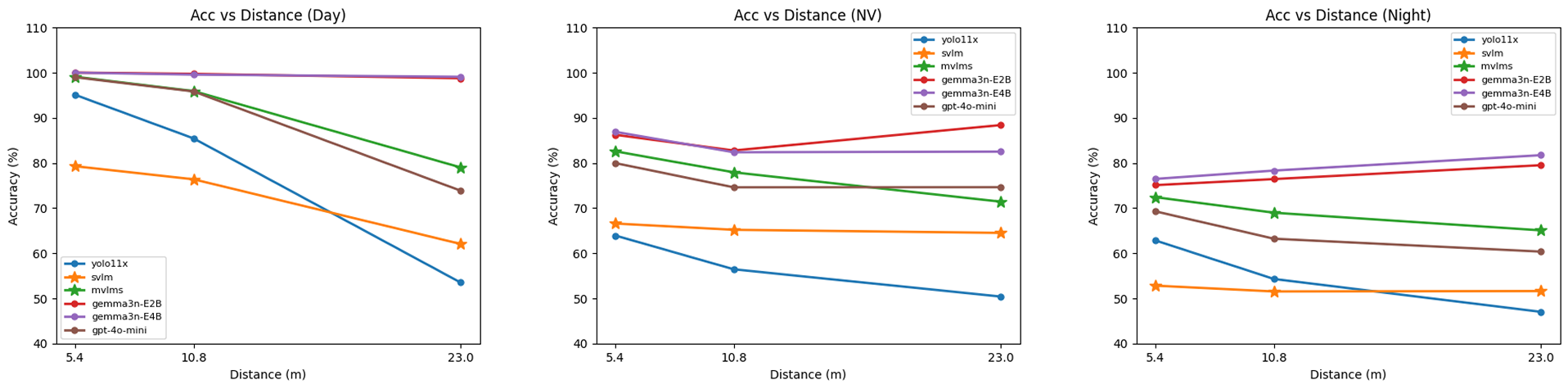}
        \caption{HPC Robustness: Accuracy across DORI-calibrated distances under different lighting conditions. The left, middle, and right subplots correspond to day, night-vision (NV), and night settings, respectively. With each subplot, distance increases from left to right, corresponding to progressively more difficult evaluation conditions.}
        \label{fig:hpc graphs}
        \end{figure}
    \end{landscape}

    To demonstrate a deeper robustness investigation via metadata stratification analysis, we analyze how degradation patterns vary with illumination by stratifying by lighting conditions. We report human-presence classification (HPC) performance over distances separately for day, night-vision, and night settings in \cref{tab:hpc distance lighting} and \cref{fig:hpc graphs}. Under \textit{day} conditions, nearly all LVLMs outperform YOLO11x, with the exception of SmolVLM2-256M-Video-Instruct and SmolVLM2-500M-Video-Instruct. This indicates that, in well-lit environments, LVLMs above roughly 1B parameters are generally capable of surpassing conventional detectors in terms of HPC and suggests that in general, LVLMs may have better perception capabilities. When inspecting how parameters affect performance, as seen in \cref{fig:hpc graphs}, YOLO and SVLMs degrade sharply at 23m, meaning that already in ideal lighting conditions, SVLMs and YOLO perform worse than our proposed failure threshold (80\%) based on reported operational performance of models \cite{mokayed2023real,liu2012vehicle,velastin2020detecting,nguyen2021yolo}. 
    
    In contrast, \textit{night} conditions lead to a substantial collapse in performance. No model attains an F1-score of 0.8, levels that many models comfortably achieve during the \textit{day}, highlighting the severe degradation induced by low-light environments. In such scenarios, all evaluated LVLMs become effectively non-operational. 
    
    Encouragingly, applying \textit{night-vision} enhancement yields partial recovery in model performance. Several models show modest but consistent gains, including gemma-3-4b-it and gemma-3n-E2B-it, suggesting that algorithmic illumination compensation is beneficial for vision-language understanding. 
    
    Across all lighting conditions and distances, the Gemma3 family: Gemma3 and Gemma3n, consistently achieves the strongest HPC performance, and specifically, gemma-3-4b-it and gemma-3n-E2B-it achieve the best overall performance. Interestingly, the smaller variants tend to outperform their larger counterparts, indicating that model scale within this family does not linearly translate to better HPC robustness. However, for the \textit{night} lighting conditions, the Mid-sized LVLMs show promise in beating gemma-3-4b-it and gemma-3n-E2B-it, this could be due to the higher reasoning capabilities of the larger models to understand the contextual clues.


\newpage

\end{document}